\documentclass[11pt]{article}

\usepackage[preprint]{acl}
\usepackage{booktabs}
\usepackage{multirow}
\usepackage{makecell}
\usepackage[table]{xcolor}
\usepackage{times}
\usepackage{latexsym}
\usepackage{amsmath}
\usepackage{float}
\usepackage{amsmath}
\usepackage{amssymb}
\usepackage{bbm}

\usepackage{algorithm}
\usepackage{algorithmic}

\usepackage[T1]{fontenc}
\usepackage[utf8]{inputenc}

\usepackage{microtype}

\usepackage{inconsolata}

\usepackage{graphicx}
\title{ConSteer-RL: Steering Reasoning Capabilities in Large Language Models \\ via Confidence-Aware Reinforcement Learning}

\author{Qing Miao$^{1}\thanks{\quad Equal Contribution}$, 
Yiming Zhao$^{2*}$, 
Jing Yang$^{1}\thanks{\quad Corresponding Author}$, 
Chenxi Liu$^{1}$, \\[2pt]
\textbf{Yuehai Chen$^{1}\footnotemark[2]$, Yuewen Liu$^{1}$, Shaoyi Du$^{1}$, Badong Chen$^{1}$} \\[2pt]
$^{1}$ Xi’an Jiaotong University \quad
$^{2}$ University of Science and Technology of China \\
}

\begin{document}
\maketitle

% % ==================插入单栏图片==================
% \begin{figure}[t]
% \centering
% \includegraphics[width=0.9\columnwidth]{figure/grpo-area-0401-v1.17_conf_distribution.png} 
% \caption{Using the trim and clip commands produces fragile layers that can result in disasters (like this one from an actual paper) when the color space is corrected or the PDF combined with others for the final proceedings. Crop your figures properly in a graphics program -- not in LaTeX..}
% \label{figure2}
% \end{figure}
% % ==================插入单栏图片==================

% % ==================插入双栏图片==================
% \begin{figure*}[t]
% \centering
% \includegraphics[width=\textwidth]{figure/grpo-area-0401-v1.17_conf_distribution.png}
% \caption{Your caption here.}
% \label{figure2}
% \end{figure*}
% % ==================插入双栏图片==================

\begin{abstract}
Reinforcement Learning from Verifiable Rewards (RLVR) has recently become a key paradigm for improving the reasoning abilities of Large Language Models (LLMs), yet it remains limited by sparse binary rewards and its ignorance of model-internal uncertainty. In this paper, we propose \textbf{ConSteer-RL}, a simple yet effective framework that integrates token-level confidence signals derived from model log-probabilities into RLVR training. Specifically, building upon the Group Relative Policy Optimization (GRPO) framework, we construct a confidence-aware reward by aggregating per-token probabilities into a scalar confidence score and incorporating it into an awareness-based reward shaping mechanism that penalizes overconfident errors while reinforcing correct and confident reasoning. Experimental results demonstrate that ConSteer-RL consistently outperforms strong GRPO baselines, achieving average improvements of 2.3\%--4.0\% across different model scales.

\end{abstract}

\section{Introduction}
RLVR has emerged as a prominent paradigm for enhancing the reasoning abilities of large language models, particularly on tasks requiring long chain-of-thought (CoT) reasoning such as mathematical reasoning~\cite{wang2026reinforcement,he2025deepmath}, code generation~\cite{jiang2025coderl+,jiang2026verifiable}, and question answering~\cite{burgess2026papersearchqa, zhao2025v2p, qi2025vcr, zeng2026vdr,zeng2025jigsaw, zhao2026videoseeker}. Unlike approaches that rely on human preference annotations, RLVR leverages rule-based or model-based verifiers to automatically assess response correctness, thereby enabling scalable, efficient, and highly automated training.

Prior work such as DeepSeek-R1~\cite{guo2025deepseek} shows that strong reasoning capabilities can emerge without explicit supervision on intermediate reasoning steps, suggesting that RLVR alone is sufficient to induce coherent COT reasoning. In parallel, advances in policy optimization algorithms have further improved the effectiveness and stability of RLVR training. In particular, GRPO~\cite{guo2025deepseek} introduces group relative advantage estimation as an alternative to learned value functions, reducing computational overhead while improving training stability and sample efficiency~\cite{shao2024deepseekmath}. These developments have significantly advanced LLM performance on reasoning tasks, establishing RLVR as a key paradigm for enhancing complex reasoning abilities and enabling scalable post-training optimization across a wide range of complex reasoning tasks.

However, existing RLVR methods still suffer from two key limitations. \textit{(1) The binary correctness reward is extremely sparse}, providing only a terminal \textit{$\pm 1$} supervision signal at the end of generation~\cite{wang2026verpo}. This makes credit assignment particularly challenging for long chain-of-thought reasoning, especially on competition-level problems where minor intermediate errors can lead to incorrect final answers~\cite{liu2026save}. \textit{(2) Most existing approaches also fail to exploit the model’s internal confidence}, treating high- and low-confidence predictions equivalently in the reward space. This design does not adequately capture the uncertainty structure of model predictions, leading to a lack of effective constraints on erroneous confidence patterns during training, and consequently limiting the model’s ability to learn reliable reasoning behaviors.

\begin{figure*}[h]
\centering
\includegraphics[width=1.0\textwidth]{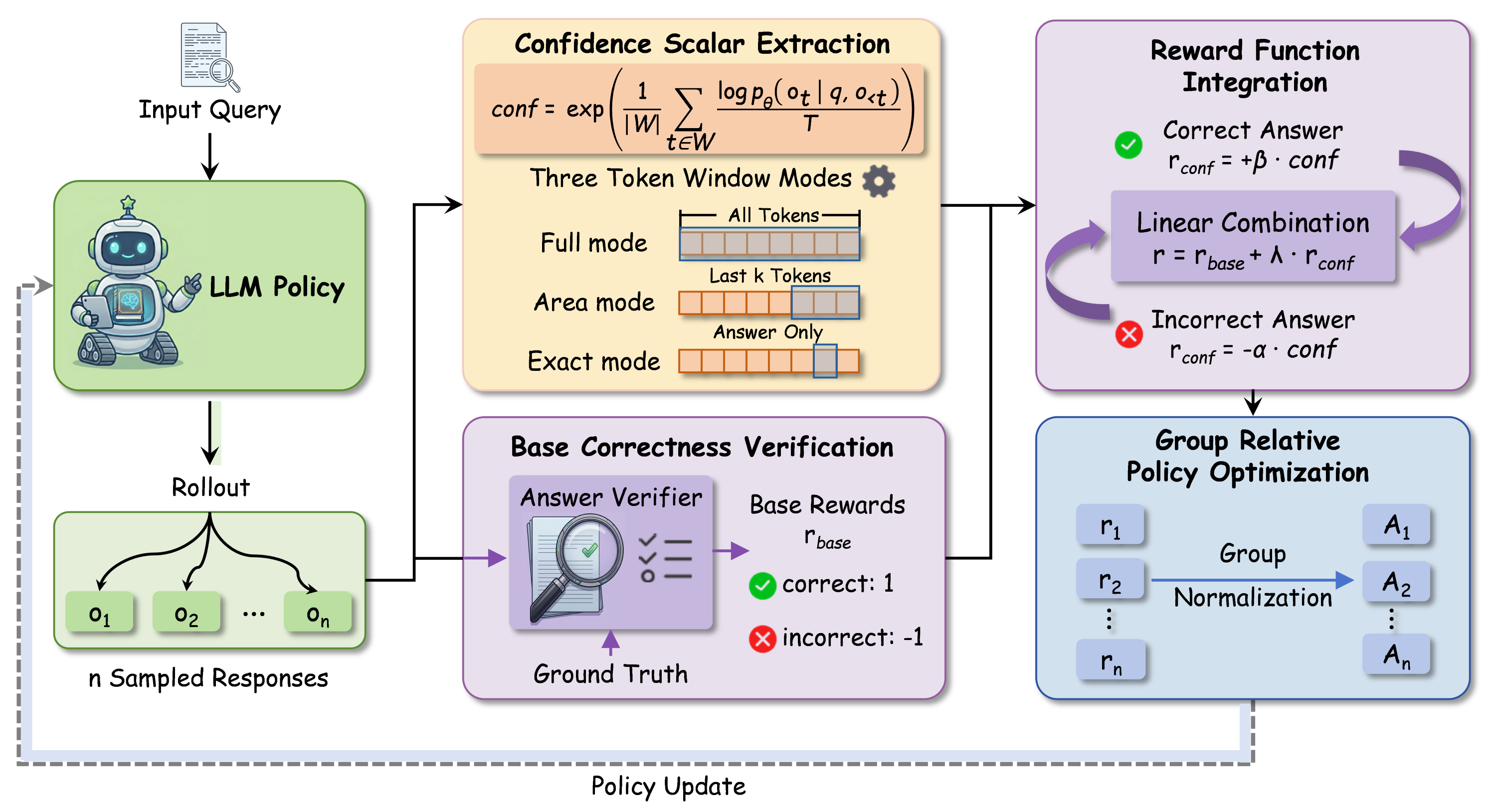}
\caption{\textbf{Overview of ConSteer-RL.} We augment RLVR by \textit{(1)} extracting a scalar confidence signal from per-token log probabilities, \textit{(2)} formulating a composite reward that integrates correctness with confidence-aware shaping, and \textit{(3)} optimizing the resulting objective with GRPO for stable policy learning.}
\label{figure:all}
\end{figure*}

To address these issues, we propose \textbf{ConSteer-RL}, a framework that integrates model-internal confidence into RLVR. As shown in Figure~\ref{figure:all}, we extract token-level confidence from log probabilities during rollouts and convert it into an additional reward component that rewards high-confidence correct responses while penalizing high-confidence incorrect ones. This confidence-aware reward is further integrated with GRPO for stable optimization. Compared with conventional RLVR methods, ConSteer-RL provides a more fine-grained and confidence-aware learning signal, encouraging more reliable reasoning behaviors. Importantly, ConSteer-RL introduces no additional reward models, auxiliary verifiers, or extra inference overhead, instead leveraging token probabilities naturally produced during rollout generation to construct an efficient confidence-aware optimization signal.

In a nutshell, our contributions are as follows:
\begin{itemize}
    \item We propose ConSteer-RL, a reinforcement learning framework that explicitly incorporates model-internal confidence signals into RLVR training, enabling simultaneous optimization of task performance and confidence-aware reasoning without requiring additional human annotations.
    
    \item We introduce a confidence-aware reward shaping mechanism that jointly models correctness and predictive uncertainty, effectively penalizing overconfident errors while reinforcing reliable reasoning trajectories.
    
    \item Extensive experiments across multiple model families and seven mathematical reasoning benchmarks demonstrate that ConSteer-RL consistently improves performance, with particularly strong gains on challenging competition-level tasks.
\end{itemize}

\section{Related Works}

\textbf{RL with Verifiable Rewards.} RLVR has become a widely adopted paradigm for improving the reasoning capabilities of large language models. Most existing methods are based on policy optimization algorithms, particularly PPO~\cite{schulman2017proximal} and GRPO~\cite{shao2024deepseekmath}. Recent works, including Dr.~GRPO~\cite{liu2025understanding}, DAPO~\cite{yu2026dapo}, GSPO~\cite{zheng2025group}, SAPO~\cite{gao2025soft}, and GDPO~\cite{liu2026gdpo} further improve training stability, reward design, and optimization objectives. Despite their effectiveness, these methods primarily optimize for final-answer correctness while largely neglecting the model’s internal confidence during reasoning, and therefore fail to fully exploit the fine-grained uncertainty signals present throughout the generation process. In contrast, our method addresses this limitation by incorporating confidence-aware rewards into the RLVR framework, enabling the training process to benefit jointly from both answer correctness and reasoning confidence constraints.

\textbf{Confidence Estimation and Optimization.} Confidence estimation quantifies the reliability of model outputs and provides complementary signals for improving LLM reasoning. Existing methods can be broadly divided into two main approaches. Black-box methods rely only on model outputs. For example, consistency-based methods estimate confidence from the agreement among multiple sampled responses~\cite{wang2022self,manakul2023selfcheckgpt,cheng2024can,yang2024alignment,han2024enhancing}, while prompting-based methods ask models to express confidence in natural language through prompts~\cite{wei2022chain,xiong2024can}. Although simple and compatible with closed-source models, these methods are limited by the model’s intrinsic self-awareness ability, often leading to unreliable confidence estimation~\cite{ji2023towards, bani2025rewarding}. White-box methods require access to internal model representations or parameters. Some approaches train probe classifiers on hidden-layer features to predict the probability of answer correctness~\cite{azaria2023internal}. Our method instead is based on the assumption that high-probability tokens correspond to high-confidence predictions~\cite{kuhn2023semantic,huang2023look,duan2024shifting}. We estimate confidence directly from token-level log probabilities during generation, without introducing additional training modules or inference overhead, making the method lightweight and efficient.

Confidence-aware optimization further utilizes confidence signals to improve model performance, and can be divided into inference-time and training-time approaches. Inference-time methods, such as DeepConf~\cite{fu2025deep}, dynamically evaluate reasoning paths during inference and improve performance by filtering low-confidence trajectories and aggregating high-confidence ones. However, these methods cannot fundamentally resolve the model’s lack of uncertainty awareness. In contrast, our method performs training-time awareness optimization by incorporating confidence-aware objectives into reinforcement learning, enabling intrinsic optimization of the model’s reasoning behavior.

\section{ConSteer-RL}
Human reasoning is naturally accompanied by self-monitoring: a solver continuously assesses their level of certainty at each step and adjusts subsequent decisions accordingly. Inspired by this metacognitive process, we propose ConSteer-RL, a framework that augments the RLVR training paradigm with an explicit, model-derived confidence signal. A scalar confidence score is first derived from per-token log probabilities produced during rollout (\S\ref{sec:confidence_extraction}). This confidence is then transformed into an additive reward component that penalizes high-confidence mistakes while rewarding self-consistent and reliable successes (\S\ref{sec:Confidence-Aware Reward Shaping}). Finally, the resulting composite reward, which integrates both task correctness and confidence signals, is optimized with Group Relative Policy Optimization to achieve stable policy updates (\S\ref{sec:Group Relative Policy Optimization}).

\subsection{Confidence Scalar Extraction}
\label{sec:confidence_extraction}
A key component of ConSteer-RL is confidence extraction, which aims to quantify the model’s certainty over its generated reasoning trajectory using a scalar confidence score. During rollout generation, vLLM provides per-token log probabilities $\log p_\theta(o_t \mid q, o_{<t})$, where $q$ denotes the input query, $o_t$ represents the $t$-th generated token, and $o_{<t}$ denotes the previously generated tokens. We aggregate these token-level probabilities into a unified confidence estimate by averaging the log probabilities over a predefined window $\mathcal{W}$ and scaling the result with a temperature hyperparameter $T$:

\begin{equation}
\mathit{conf}
= \exp\!\left(
    \frac{1}{|\mathcal{W}|}
    \sum_{t \in \mathcal{W}}
    \frac{\log p_\theta(o_t \mid q, o_{<t})}{T}
\right)
\end{equation}

\begin{figure}[t]
\centering
\includegraphics[width=0.95\columnwidth]{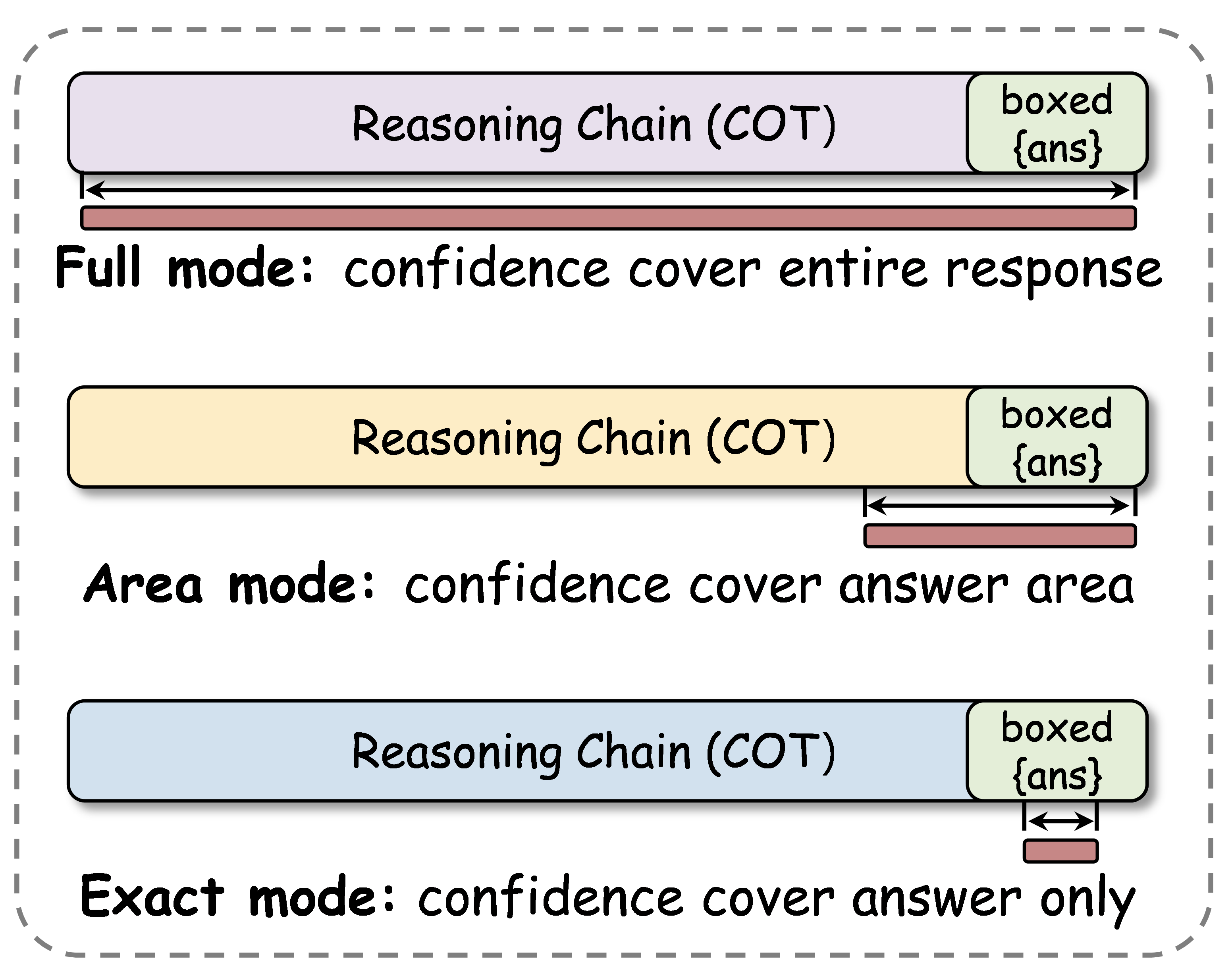} 
\caption{Definitions of token-window modes.}
\label{figure:mode}
\end{figure}

As shown in Figure~\ref{figure:mode}, we design three token-window modes to define $\mathcal{W}$:
\begin{itemize}
\item \textbf{Full Mode}: $\mathcal{W}$ covers the entire generated response, providing a global estimate of the model’s overall confidence throughout the reasoning process.

\item \textbf{Area Mode}: $\mathcal{W}$ is restricted to the last $K$ generated tokens, emphasizing the final answer composition stage while filtering out intermediate scratchpad noise.

\item \textbf{Exact Mode}: For tasks containing explicitly formatted boxed answers (e.g., \texttt{\textbackslash boxed\{...\}}), $\mathcal{W}$ is the token span inside the boxed region, enabling confidence estimation to focus directly on the predicted final answer. It falls back to area mode if no box is found.
\end{itemize}

All three modes share the same formulation, differing only in the choice of token window $\mathcal{W}$.

\subsection{Reward Function Integration}
\label{sec:Confidence-Aware Reward Shaping}
Equipped with the confidence score $\mathit{conf}$, we propose a confidence-aware reward shaping mechanism that augments the sparse binary correctness signal with an additional reward dimension, thereby yielding a more informative learning signal. Let $a$ denote the model prediction and $a^\star$ the ground-truth. A rule-based answer verifier is first applied to determine whether the model output matches the ground truth, producing a base reward signal:

\begin{equation}
r_{\textit{base}} = \mathbbm{1}[a = a^\star] - \mathbbm{1}[a \neq a^\star]
\end{equation}

We then compute a confidence reward $r_{\text{conf}}$ as:
% \begin{equation}
% r_{\textit{conf}} =
% \begin{cases}
% -\alpha \cdot \mathit{conf} & \text{if answer is incorrect} \\
% +\beta \cdot \mathit{conf} & \text{if answer is correct}
% \end{cases}
% \end{equation}

\begin{equation}
r_{\textit{conf}} =
\beta \, \mathit{conf} \cdot \mathbbm{1}[a = a^\star]
-
\alpha \, \mathit{conf} \cdot \mathbbm{1}[a \neq a^\star]
\end{equation}

The penalty coefficient $\alpha$ controls the strength of the penalty applied when the model produces an incorrect answer with high confidence; larger values of $\alpha$ impose stronger penalties on such overconfident errors, thereby discouraging incorrect yet high-confidence predictions. The bonus coefficient $\beta$ determines the additional reward assigned to correct answers with high confidence; increasing $\beta$ places greater emphasis on reinforcing reliable and consistent reasoning trajectories, encouraging the model to remain both accurate and confidently assertive when making correct predictions.

The final reward is defined as a linear combination of the base correctness score and a confidence shaping term:
\begin{equation}
r = r_{\textit{base}} + \lambda \cdot r_{\textit{conf}}
\end{equation}
where $\lambda > 0$ controls the overall contribution of the confidence signal. This formulation integrates predictive confidence with task correctness into a unified learning objective, enabling the policy to jointly optimize task performance and uncertainty handling in a principled manner.

\subsection{Group Relative Policy Optimization}
\label{sec:Group Relative Policy Optimization}

We adopt Group Relative Policy Optimization~\cite{shao2024deepseekmath} as our policy optimization algorithm. For each query $q$, the current policy $\pi_\theta$ independently samples $n$ responses $\{o_1, \ldots, o_n\}$. GRPO dispenses with a learned value function and uses the within-group reward statistics as a baseline instead, making it more sample-efficient.

Each response $o_i$ receives the confidence-aware composite reward $\tilde{r}_i = r_{\textit{base},i} + \lambda \cdot r_{\textit{conf},i}$ (defined in \S\ref{sec:Confidence-Aware Reward Shaping}), where $r_{\textit{base},i} \in \{+1, -1\}$ and $r_{\textit{conf},i}$ is the confidence-dependent shaping term. GRPO normalizes these rewards within each query group to obtain stable advantages. For the group $\{\tilde{r}_j\}_{j=1}^n$ associated with query $q$, let $\mu_q$ and $\sigma_q$ denote the group mean and standard deviation:
\begin{equation}
\mu_q = \frac{1}{n} \sum_{j=1}^{n} \tilde{r}_j \qquad
\sigma_q = \sqrt{\frac{1}{n} \sum_{j=1}^{n} (\tilde{r}_j - \mu_q)^2 + \delta}
\end{equation}
where $\delta > 0$ is a small constant for numerical stability. The normalized advantage for response $o_i$ is:
\begin{equation}
\hat{A}_i = \frac{\tilde{r}_i - \mu_q}{\sigma_q}
\end{equation}
This normalization standardizes rewards within each group, stabilizing gradients and reducing the impact of varying question difficulty.

The policy is updated by maximizing the following policy gradient objective:
\begin{equation}
\label{eq:grpo_loss}
\begin{aligned}
\mathcal{L}_{\text{GRPO}}(\theta)
&= \frac{1}{G} \sum_{q=1}^{G} \frac{1}{n} \sum_{i=1}^{n}
   \frac{1}{|o_i|} \sum_{t=1}^{|o_i|}
   \min \Bigl(
   r_{i,t}(\theta)\hat{A}_i, \\
&\qquad
   \operatorname{clip}\bigl(r_{i,t}(\theta), 1-\epsilon, 1+\epsilon\bigr)\hat{A}_i
   \Bigr)
\end{aligned}
\end{equation}

where the importance sampling ratio is defined as
\begin{equation}
r_{i,t}(\theta)
= \frac{\pi_\theta(o_{i,t} \mid q, o_{i,<t})}
{\pi_{\theta_{\text{old}}}(o_{i,t} \mid q, o_{i,<t})}
\end{equation}

Here, $G$ denotes the number of queries, $n$ is the number of sampled responses per query, and $|o_i|$ is the length of response $o_i$, $\epsilon$ specifies the clipping range to ensure conservative policy updates. The term $\pi_\theta(o_{i,t} \mid q, o_{i,<t})$ represents the probability of generating token $o_{i,t}$ conditioned on the query $q$ and preceding tokens $o_{i,<t}$.

Confidence-aware rewards combined with advantage normalization form a closed optimization loop. High-confidence but incorrect responses receive stronger penalties, which are then reflected through negative advantage signals that suppress similar trajectories in subsequent updates. Conversely, high-confidence correct responses are assigned positive advantages, reinforcing their corresponding reasoning paths and increasing their likelihood in future sampling.

To better illustrate the overall training pipeline of ConSteer-RL, the complete procedure is summarized in Algorithm~\ref{alg:consteer}.

\begin{algorithm}[t]
\footnotesize
\caption{ConSteer-RL Training Procedure}
\label{alg:consteer}

\begin{algorithmic}[1]

\REQUIRE Training set $\mathcal{D}$, policy $\pi_\theta$, group size $n$, temperature $T$, coefficients $\alpha,\beta,\lambda$.
\ENSURE Updated policy parameters $\theta$.

\FOR{each training iteration.}

    \STATE Sample query batch $\{q_1,\dots,q_G\} \sim \mathcal{D}$.  

    \FOR{each query $q$}

        \STATE Sample $n$ responses $\{o_i\}_{i=1}^{n} \sim \pi_\theta(\cdot \mid q)$.  

        \STATE Extract confidence $\mathit{conf}_i$ using window mode $\mathcal{W}_i$.  

        \STATE Compute rewards $(r_{\textit{base},i}, r_{\textit{conf},i}, \tilde r_i)$.  

        \STATE Normalize group rewards $\{\tilde r_i\}_{i=1}^n$ into advantages $\{\hat A_i\}_{i=1}^n$.  

    \ENDFOR

    \STATE Update policy $\pi_\theta$ via GRPO using advantages $\hat A_i$.  

\ENDFOR

\RETURN $\theta$

\end{algorithmic}
\end{algorithm}

\section{Experiments}
\subsection{Experimental Setups}

\textbf{Models and Datasets.} To demonstrate the general applicability of ConSteer-RL, we conduct experiments across a diverse range of model scales including Qwen2.5-Math-7B~\cite{yang2024qwen2}, Qwen3-4B-Base, Qwen3-8B-Base~\cite{yang2025qwen3}, performing direct reinforcement learning using the GRPO framework. We train our models on the DAPO-Math-17k dataset~\cite{yu2025dapo}, which contains diverse problems spanning multiple difficulty levels. For evaluation, we consider seven challenging benchmarks, including MATH500~\cite{hendrycks2020measuring}, Minerva Math~\cite{lewkowycz2022solving}, OlympiadBench~\cite{he2024olympiadbench}, as well as AIME24, AIME25, AIME26, and AMC23~\cite{he2024olympiadbench}. These benchmarks cover a broad spectrum of difficulty, providing a comprehensive assessment of the models’ mathematical reasoning ability and generalization performance.

\textbf{Implementation Details.} All reinforcement learning experiments are conducted using verl~\cite{sheng2025hybridflow} on 8 $\times$ A100 GPUs. We employ math-verify as the verifier during training and simpleRL verifier~\cite{zeng2025simplerl} for final evaluation. During training, we sample 8 responses per prompt at a temperature of 1.0 with a batch size of 128. The model is optimized with a learning rate of $1 \times 10^{-6}$ and a mini-batch size of 64. For evaluation, we report Avg@1 at temperature $0$ for Math500, Minerva Math, and OlympiadBench, and Avg@32 at temperature $0.7$ with top\_p set to $0.8$ for AIME24/25/26 and AMC23. No confidence-based interventions are applied during inference to ensure a clean evaluation setting. To prevent potential reward hacking, where a group of entirely incorrect trajectories may nevertheless receive spurious positive rewards due to lower confidence, we incorporate the Dynamic Sampling mechanism from DAPO~\cite{yu2026dapo}. Further details are provided in Appendix~\ref{sec:Hyperparameters}.

\subsection{Experimental Results}

\begin{table*}[t]
    \centering
    \caption{\textbf{Main Results Comparison.} We report the performance of three models across seven mathematical reasoning benchmarks. ConSteer-RL achieves the best overall performance across all model scales, consistently outperforming both the Baseline and GRPO. The best results are highlighted in \textbf{bold}.}
    \label{tab:main_results}

    \setlength{\tabcolsep}{4.5pt}
    \fontsize{10.0pt}{11.5pt}\selectfont

    \resizebox{\textwidth}{!}{
    \begin{tabular}{p{1.9cm} c c c c c c c c}
        \toprule
        \textbf{Model} & \textbf{Math500} & \makecell{\textbf{Minerva}\\\textbf{math}} & \makecell{\textbf{Olympiad}\\\textbf{Bench}} & \textbf{AIME24} & \textbf{AIME25} & \textbf{AIME26} & \textbf{AMC23} & \textbf{Avg.} \\
        \midrule
        \multicolumn{9}{l}{\textit{\textbf{Qwen2.5-Math-7B}}} \\
        Baseline & 65.0 & 17.6 & 25.2 & 16.7 & 6.7  & 6.7  & 35.0 & 24.7 \\
        GRPO     & 77.6 & 34.9 & 37.9 & 23.3 & 10.0 & 10.0 & 60.0 & 36.2 \\
        \rowcolor{blue!5}
        ConSteer-RL  & \textbf{78.4} & \textbf{40.1} & \textbf{38.5} & \textbf{26.7} & \textbf{13.3} & \textbf{16.7} & \textbf{67.5} & \textbf{40.2} \\
        \textcolor{green!60!black}{$\Delta_{\text{GRPO}}$} &
        \textbf{\textcolor{green!60!black}{+0.8}} & 
        \textbf{\textcolor{green!60!black}{+5.2}} &
        \textbf{\textcolor{green!60!black}{+0.6}} &
        \textbf{\textcolor{green!60!black}{+3.4}} &
        \textbf{\textcolor{green!60!black}{+3.3}} &
        \textbf{\textcolor{green!60!black}{+6.7}} &
        \textbf{\textcolor{green!60!black}{+7.5}} &
        \textbf{\textcolor{green!60!black}{+4.0}} \\
        \midrule

        \multicolumn{9}{l}{\textit{\textbf{Qwen3-4B-Base}}} \\
        Baseline & 71.4 & 26.5 & 36.1 & 12.0 & 6.8  & 10.7 & 41.6 & 29.3 \\
        GRPO     & 83.6 & 42.6 & 46.7 & 23.2 & 23.4 & 14.0 & 63.0 & 42.4 \\
        \rowcolor{blue!5}
        ConSteer-RL  & \textbf{85.8} & \textbf{45.2} & \textbf{47.0} & \textbf{26.5} & \textbf{26.6} & \textbf{20.2} & \textbf{63.8} & \textbf{45.0} \\
        \textcolor{green!60!black}{$\Delta_{\text{GRPO}}$} &
        \textbf{\textcolor{green!60!black}{+2.2}} &
        \textbf{\textcolor{green!60!black}{+2.6}} &
        \textbf{\textcolor{green!60!black}{+0.3}} &
        \textbf{\textcolor{green!60!black}{+3.3}} &
        \textbf{\textcolor{green!60!black}{+3.2}} &
        \textbf{\textcolor{green!60!black}{+6.2}} &
        \textbf{\textcolor{green!60!black}{+0.8}} &
        \textbf{\textcolor{green!60!black}{+2.6}} \\
        \midrule

        \multicolumn{9}{l}{\textit{\textbf{Qwen3-8B-Base}}} \\
        Baseline & 76.6 & 37.9 & 40.6 & 10.2 & 13.3 & 6.2  & 51.7 & 33.8 \\
        GRPO     & 86.0 & 46.7 & 51.7 & 25.4 & 24.9 & 20.7 & 69.2 & 46.4 \\
        \rowcolor{blue!5}
        ConSteer-RL  & \textbf{86.6} & \textbf{49.3} & \textbf{53.9} & \textbf{26.6} & \textbf{29.8} & \textbf{23.8} & \textbf{70.9} & \textbf{48.7} \\
        \textcolor{green!60!black}{$\Delta_{\text{GRPO}}$} &
        \textbf{\textcolor{green!60!black}{+0.6}} &
        \textbf{\textcolor{green!60!black}{+2.6}} &
        \textbf{\textcolor{green!60!black}{+2.2}} &
        \textbf{\textcolor{green!60!black}{+1.2}} &
        \textbf{\textcolor{green!60!black}{+4.9}} &
        \textbf{\textcolor{green!60!black}{+3.1}} &
        \textbf{\textcolor{green!60!black}{+1.7}} &
        \textbf{\textcolor{green!60!black}{+2.3}} \\
        \bottomrule
    \end{tabular}
    }

\end{table*}

Table~\ref{tab:main_results} presents the main experimental results, from which we draw several key observations. First, ConSteer-RL consistently outperforms GRPO across all benchmarks and model families, demonstrating strong robustness across scales. Second, the performance improvements are stable and non-trivial, with average gains of 4.0\% for Qwen2.5-Math-7B, 2.6\% for Qwen3-4B-Base, and 2.3\% for Qwen3-8B-Base. Third, the improvements are particularly pronounced on challenging competition-level benchmarks, such as the AIME series. This can be attributed to the intrinsic difficulty of such tasks, where long-horizon reasoning exacerbates the sparsity of binary reward signals, limiting their effectiveness for providing meaningful supervision at intermediate reasoning steps. In contrast, our confidence-based signal provides denser and more fine-grained feedback, which more effectively guides policy optimization over complex intermediate reasoning trajectories.

\begin{figure}[t]
\centering
\includegraphics[width=1.0\columnwidth]{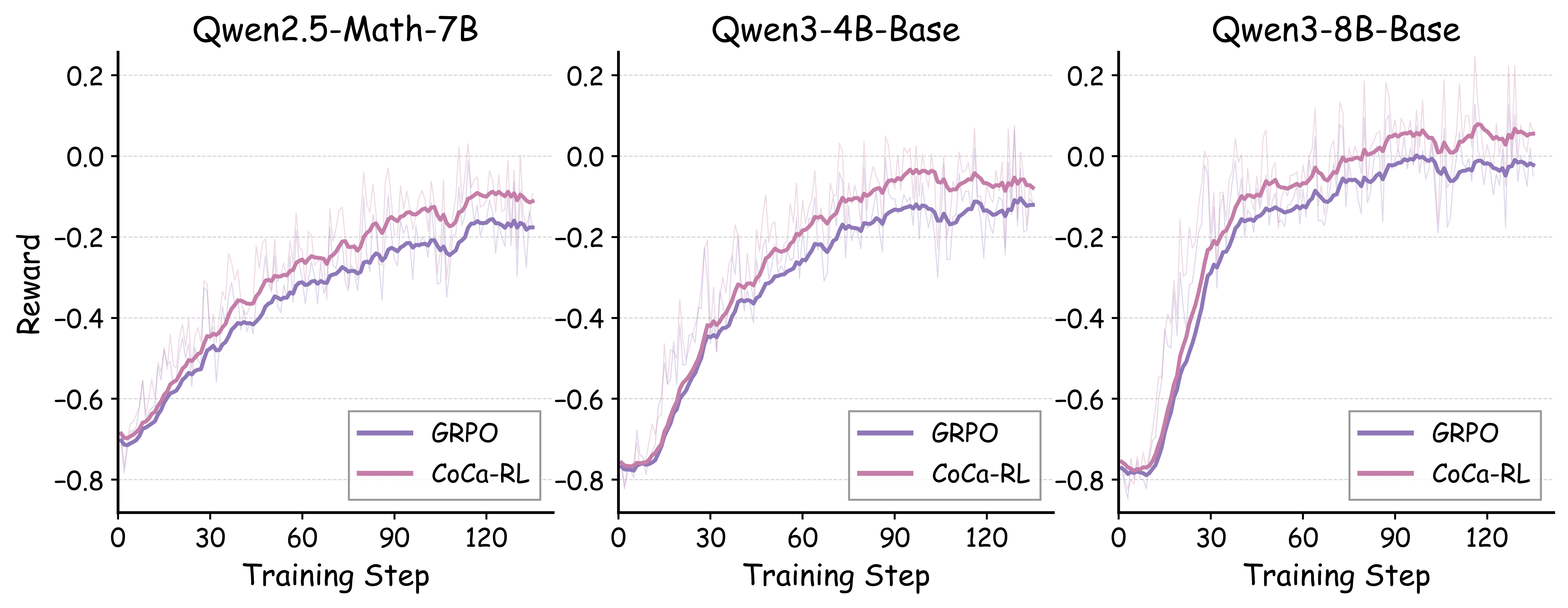} 
\caption{Comparison of training reward curves between GRPO and ConSteer-RL. ConSteer-RL rewards are rescaled to align with the range of GRPO for visualization purposes.}
\label{figure2}
\end{figure}

\begin{table}[t]
\centering
\caption{Generalization evaluation of ConSteer-RL under different training method settings.The best results are highlighted in \textbf{bold}.}
\resizebox{\columnwidth}{!}{
\begin{tabular}{lccccc}
\toprule
\textbf{Method} & \textbf{AIME24} & \textbf{AIME25} & \textbf{AIME26} & \textbf{AMC23} & \textbf{Avg.} \\
\midrule
GSPO            & 24.1 & 11.1 &  9.4 & 63.3 & 27.0 \\
ConSteer-RL     & \textbf{27.1} & \textbf{13.8} & \textbf{10.4} & \textbf{66.7} & \textbf{29.5} \\
\midrule
DAPO            & 25.5 & 12.3 & 11.5 & 65.4 & 28.7 \\
ConSteer-RL  & \textbf{29.2} & \textbf{15.6} & \textbf{14.5} & \textbf{68.2} & \textbf{31.9} \\
\midrule
SAPO            & 24.6 & 10.7 & 10.2 & 61.2 & 26.7 \\
ConSteer-RL     & \textbf{27.7} & \textbf{11.5} & \textbf{12.8} & \textbf{63.1} & \textbf{28.8} \\
\bottomrule
\end{tabular}
}
\label{tab:对比results}
\end{table}

As illustrated in Figure~\ref{figure2}, ConSteer-RL consistently outperforms the vanilla GRPO baseline across all three model scales. The reward curve rises more steeply in the early stage of training, indicating faster initial learning, and maintains a persistent advantage throughout the entire optimization process. Notably, the performance gap continues to widen in later stages, ultimately converging to a substantially higher reward plateau. This pattern suggests that the improved confidence-aware reward modeling in ConSteer-RL provides stronger and more stable optimization signals, which not only accelerates early-stage learning but also enables more effective exploration in the later stages of training, leading to stronger final performance.

To further validate the generalizability of the ConSteer-RL method, we extend its application to GSPO~\cite{zheng2025group}, DAPO~\cite{yu2026dapo}, and SAPO~\cite{gao2025soft} built upon the Qwen2.5-Math-7B backbone, and conduct experiments under identical settings for fair comparison. As shown in Table~\ref{tab:对比results}, incorporating ConSteer-RL consistently leads to stable and uniform performance gains across all three baseline algorithms. This demonstrates that the proposed method is not only compatible with different RL optimization frameworks but also exhibits strong transferability and broad applicability across algorithmic variants.

\subsection{Ablation Studies}
To better understand the impact of key design choices in ConSteer-RL, we conduct ablation studies on the temperature scaling factor $T$ and the token-window mode. For narrative clarity, this section grounds the detailed analysis in the Qwen2.5-Math-7B model, results for the Qwen3 series exhibit consistent trends and are detailed in Appendix~\ref{Temperature Scaling Factor Ablation} and~\ref{Token-Window Mode Ablation}.

\textbf{Effect of Temperature Scaling $T$.} 
The temperature scaling factor $T$ rescales token log probabilities in confidence computation, thereby affecting the resulting confidence distribution. To isolate this effect, all variants are evaluated under the Full token-window mode. As reported in Table~\ref{tab:ablation_t-zhengwen}, $T=0.8$ yields the best performance.

\begin{figure}[t]
\centering
\includegraphics[width=1.0\columnwidth]{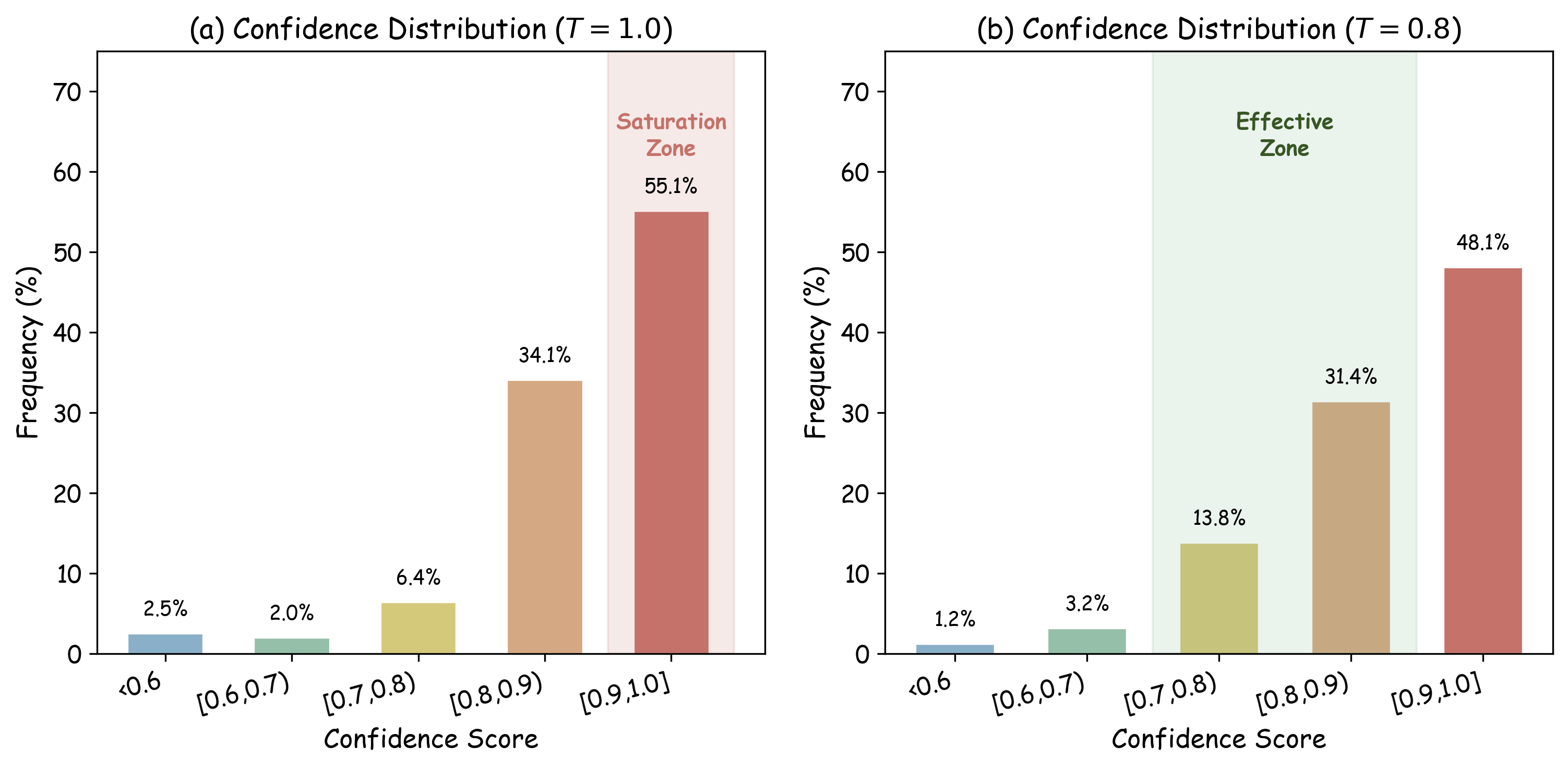} 
\caption{Effect of temperature scaling $T$ on the confidence distribution. Lowering $T$ effectively dispersing and deconcentrating the confidence mass.}
\label{fig:temp_analysis}
\end{figure}

We further examine this effect through empirical confidence distributions shown in Figure~\ref{fig:temp_analysis}. Under the default setting $T = 1.0$, pretrained models tend to assign consistently high probabilities to generated tokens. As a result, the resulting confidence scores become heavily concentrated within a narrow high-probability region, as illustrated in Figure~\ref{fig:temp_analysis} a, leading to a pronounced saturation effect. In this regime, many samples receive similarly high confidence values, which reduces the effective separability among different reasoning trajectories. In contrast, lowering the temperature to $T = 0.8$ redistributes the confidence mass away from the saturated high-confidence region, as shown in Figure~\ref{fig:temp_analysis} b, producing a noticeably more dispersed and structured distribution. This rescaling alleviates the saturation effect observed under $T = 1.0$ and expands the effective confidence range across samples. Consequently, confidence scores become more informative and better differentiated, improving the separability among reasoning trajectories. This leads to more discriminative reward signals across samples and ultimately provides a stronger optimization signal during learning.

\begin{table}[t]
    \centering
    \caption{Ablation on Temperature Scaling}
    \label{tab:ablation_t-zhengwen}
    \small
    \begin{tabular}{lccc}
        \toprule
        \textbf{Benchmark} & $\mathbf{T=1.0}$ & $\mathbf{T=0.8}$ & $\mathbf{T=0.5}$ \\
        \midrule
        Math500 & 75.0 & \textbf{78.4} & 75.4 \\
        Minerva Math & 41.2 & 40.1 & \textbf{41.5} \\
        OlympiadBench & \textbf{38.7} & 38.5 & 37.9 \\
        AIME24 & 20.0 & \textbf{26.7} & 20.0 \\
        AIME25 & 10.0 & \textbf{13.3} & 6.7 \\
        AIME26 & 10.0 & \textbf{16.7} & 6.7 \\
        AMC23 & \textbf{67.5} & \textbf{67.5} & \textbf{67.5} \\
        \midrule
        \textbf{Average} & 37.5 & \textbf{40.2} & 36.5 \\
        \bottomrule
    \end{tabular}
\end{table}

\textbf{Effect of Token-Window Mode.} We also ablate the token window $\mathcal{W}$ used for confidence calculation. We compare the three modes under the same experimental setting. To isolate the effect of the mode, all variants are evaluated at $T=1.0$. As demonstrated in Table~\ref{tab:ablation_mode-zhengwen}, the Full mode consistently achieves superior performance over both restricted alternatives. 

\begin{figure}[t]
\centering
\includegraphics[width=1.0\columnwidth]{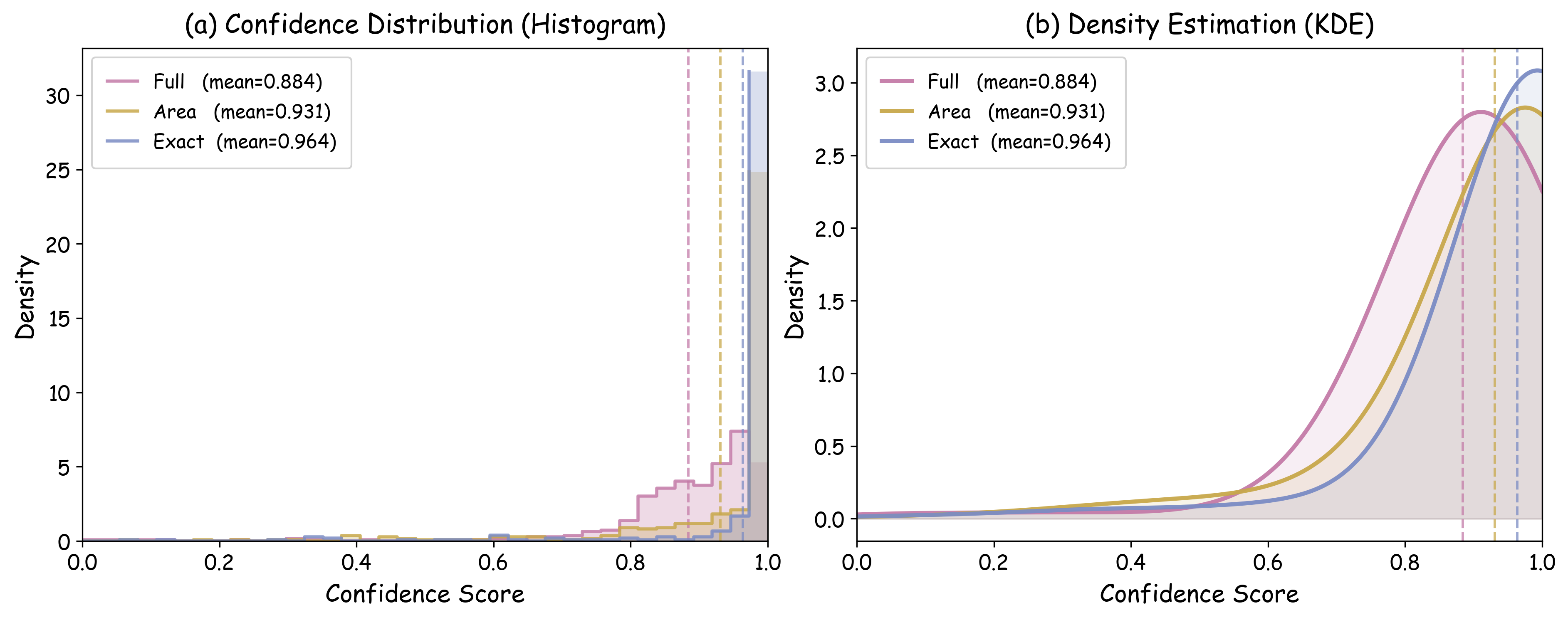} 
\caption{Analysis of token-window modes on empirical confidence distributions. Restricted modes exhibit severe confidence collapse, whereas the Full mode maintains a broader and more balanced distribution.}
\label{fig:mode_analysis}
\end{figure}

Narrowing the window induces what we term confidence collapse. This phenomenon arises because the model is inherently overconfident on standard structural or formatting tokens that typically appear near the end of responses. When confidence is computed exclusively over such a restricted subset of tokens, the resulting scores become systematically inflated and thus fail to faithfully reflect the quality of the underlying reasoning process. As illustrated in Figure~\ref{fig:mode_analysis}, the confidence distributions under the restricted modes (including the Area and Exact modes) are therefore heavily concentrated near $1.0$, exhibiting a pronounced saturation effect. This distributional collapse further creates an evaluation bias, where high-confidence formatting tokens disproportionately dominate the aggregated score. Consequently, the model can trivially hack the reward by simply generating highly confident formatting tokens, allowing it to maximize the objective without improving the actual quality of reasoning. This ultimately leads to a degeneration of the training objective into memorization and exploitation of superficial output formats, rather than fostering meaningful improvements in fundamental reasoning capabilities.

\begin{table}[t]
    \centering
    \caption{Ablation on Token-Window Mode}
    \label{tab:ablation_mode-zhengwen}
    \small
    \begin{tabular}{lccc}
        \toprule
        \textbf{Benchmark} & \textbf{Full} & \textbf{Area} & \textbf{Exact} \\
        \midrule
        Math500 & 75.0 & 74.2 & \textbf{76.4} \\
        Minerva Math & 41.2 & \textbf{41.5} & 39.0 \\
        OlympiadBench & \textbf{38.7} & 37.8 & 37.3 \\
        AIME24 & 20.0 & 22.2 & \textbf{24.7} \\
        AIME25 & \textbf{10.0} & \textbf{10.0} & 5.1 \\
        AIME26 & 10.0 & \textbf{13.3} & \textbf{13.3} \\
        AMC23 & \textbf{67.5} & 59.8 & 57.6 \\
        \midrule
        \textbf{Average} & \textbf{37.5} & 37.0 & 36.2 \\
        \bottomrule
    \end{tabular}
\end{table}

\begin{figure*}[t]
\centering
\includegraphics[width=0.94\textwidth]{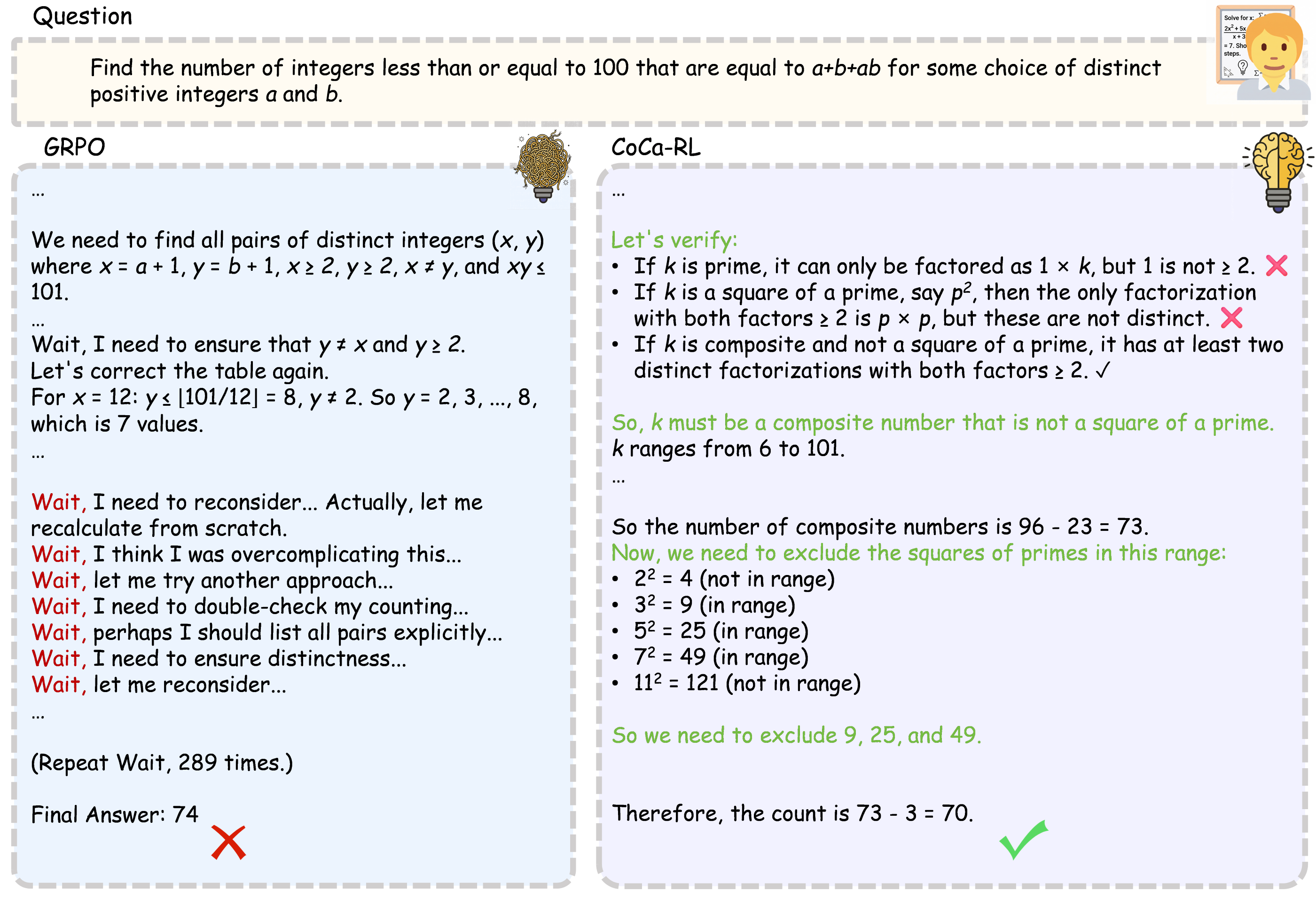}
\caption{Case Study. GRPO falls into a 289 step “Wait” loop and fails to recover; ConSteer-RL steadily applies factorization and identifies the key non-square constraint, yielding the correct answer.}
\label{figure:case-main}
\end{figure*}

\subsection{How Confidence-Aware Reward Shapes Reasoning Patterns}

To further investigate how the confidence-aware reward shapes reasoning behaviors beyond quantitative metrics, we conduct a qualitative comparison of reasoning trajectories between ConSteer-RL and the GRPO baseline. Through representative case studies, we analyze how confidence signals regulate reasoning depth and influence the model’s decision-making behavior and reasoning strategy under different levels of uncertainty.

As shown in Figure~\ref{figure:case-main}, confidence should not be interpreted as blind certainty, but as a signal reflecting the model’s internal assessment of its own reasoning quality and the reliability of intermediate conclusions. In contrast, GRPO tends to fall into cycles of low-confidence self-doubt, whereas ConSteer-RL exhibits more structured and adaptive reasoning behavior: when confidence is high, the model proceeds more decisively toward the final answer, avoiding unnecessary redundant steps and over-deliberation; when confidence is low, it correspondingly extends its reasoning chain to support more thorough exploration, cross-checking, and verification of intermediate results.

Overall, ConSteer-RL achieves a better balance between decisiveness under high confidence and cautious exploration under uncertainty. This balance not only results in shorter and more efficient reasoning chains, but also significantly improves answer accuracy and enhances the overall stability of model performance.

\section{Conclusion}
We propose ConSteer-RL, a confidence-aware RL framework that enhances RLVR by incorporating model-intrinsic uncertainty into reward design. Using token-level log-probability statistics, it derives a confidence signal that complements sparse correctness rewards and enables more fine-grained learning. Integrated with GRPO, it mitigates overconfident failure modes while strengthening reliable reasoning trajectories. Experiments across multiple benchmarks and model scales show that ConSteer-RL consistently outperforms RL baselines. Further analysis highlights the importance of temperature scaling and full token-window mode in alleviating confidence saturation and mitigating reward hacking behaviors. Overall, our results underscore the value of uncertainty-aware reward design for more reliable language models. We hope that ConSteer-RL provides the community with a scalable paradigm and fosters the development of more robust RLVR methods as well as more trustworthy language model systems.

\section*{Limitations}

The current evaluation is primarily conducted on mathematical tasks, which provide a clean and controlled testbed for validating the proposed method. However, such a setting may not fully capture its behavior in broader application scenarios, especially when facing more heterogeneous inputs, ambiguous instructions, or noisy real-world contexts. In future work, we plan to extend our evaluation to a wider range of domains, including general STEM problems and multimodal tasks, to further assess the generality and robustness of the proposed approach, and to better understand how it transfers beyond strictly structured reasoning benchmarks.

% \input{section/8_ack}

% 参考文献
\bibliography{custom}

\clearpage
\appendix

\section*{Appendix Overview}
\begin{itemize}
    \setlength{\itemsep}{0pt}
    \setlength{\topsep}{0pt}
    \setlength{\parsep}{0pt}
    \setlength{\partopsep}{0pt}
    \item Section~\ref{sec:appendix_ablation}: Extended Ablation Studies.
    \item Section~\ref{sec:Hyperparameters}: Hyperparameters.
    \item Section~\ref{sec:Case Study}: Case Studies.
    \item Section~\ref{sec:Radar Chart Analysis Across Different Models}: Analysis Across Models.
    \item Section~\ref{sec:Dataset Details}: Dataset Details.
    \item Section~\ref{sec:Benchmark Information}: Benchmark Information.
\end{itemize}

\section{Extended Ablation Studies}
\label{sec:appendix_ablation}
\subsection{Temperature Scaling Factor Ablation}
\label{Temperature Scaling Factor Ablation}

\begin{figure}[!htbp]
\centering
\includegraphics[width=1.0\columnwidth]{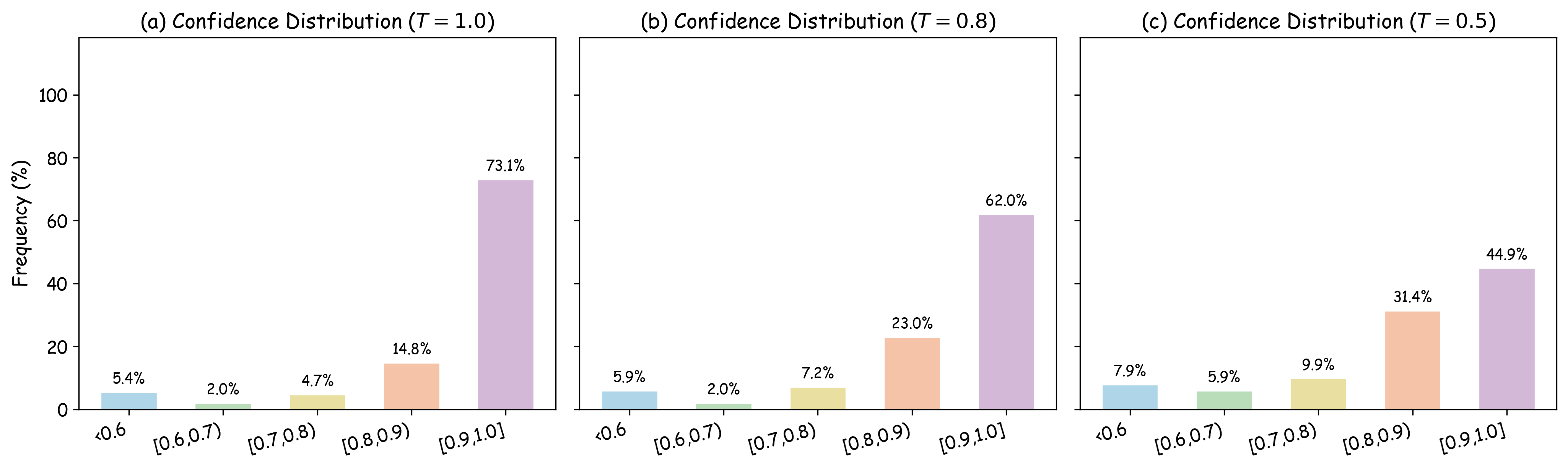} 
\caption{Confidence distribution of Qwen3-4B-Base across different temperatures $T$.}
\label{fig:qwen3-4b-appendix-T}
\end{figure}

\begin{figure}[t]
\centering
\includegraphics[width=1.0\columnwidth]{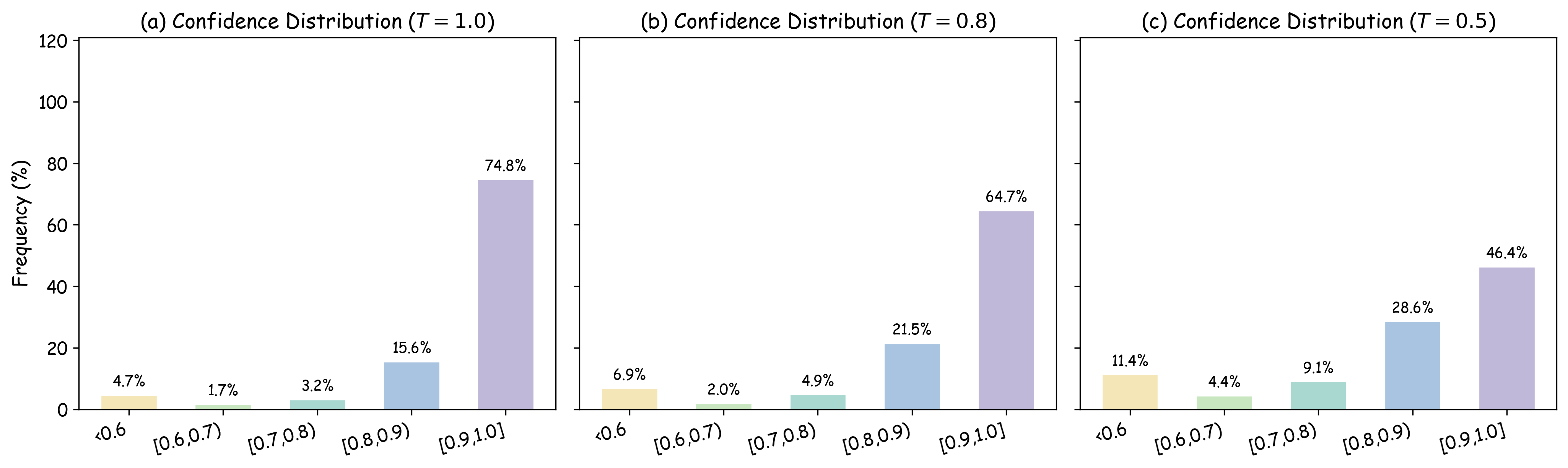} 
\caption{Confidence distribution of Qwen3-8B-Base across different temperatures $T$.}
\label{fig:qwen3-8b-appendix-T}
\end{figure}

As shown in Table~\ref{tab:ablation_t-appendix}, the Qwen3 models achieve optimal performance when the temperature is set to $T=0.5$. In contrast to Qwen2.5, the Qwen3 base models exhibit a pronounced \emph{confidence saturation} phenomenon under the default temperature $T=1.0$. Specifically, as illustrated in Figures~\ref{fig:qwen3-4b-appendix-T} and \ref{fig:qwen3-8b-appendix-T}, more than 70\% of samples fall within the high-confidence interval $[0.9, 1.0]$. Such an overly peaked distribution substantially reduces the variance of the reward signal, causing the reward gradients to nearly vanish and thereby undermining the effectiveness of optimization. When the temperature is reduced to $T=0.5$, this issue is significantly alleviated, with fewer than 50\% of samples remaining in the $[0.9, 1.0]$ range. The lower temperature effectively spreads the confidence distribution, increasing the discriminative resolution between samples and providing a more stable and informative learning signal for the optimizer.

\begin{figure}[t]
\centering
\includegraphics[width=0.7\columnwidth]{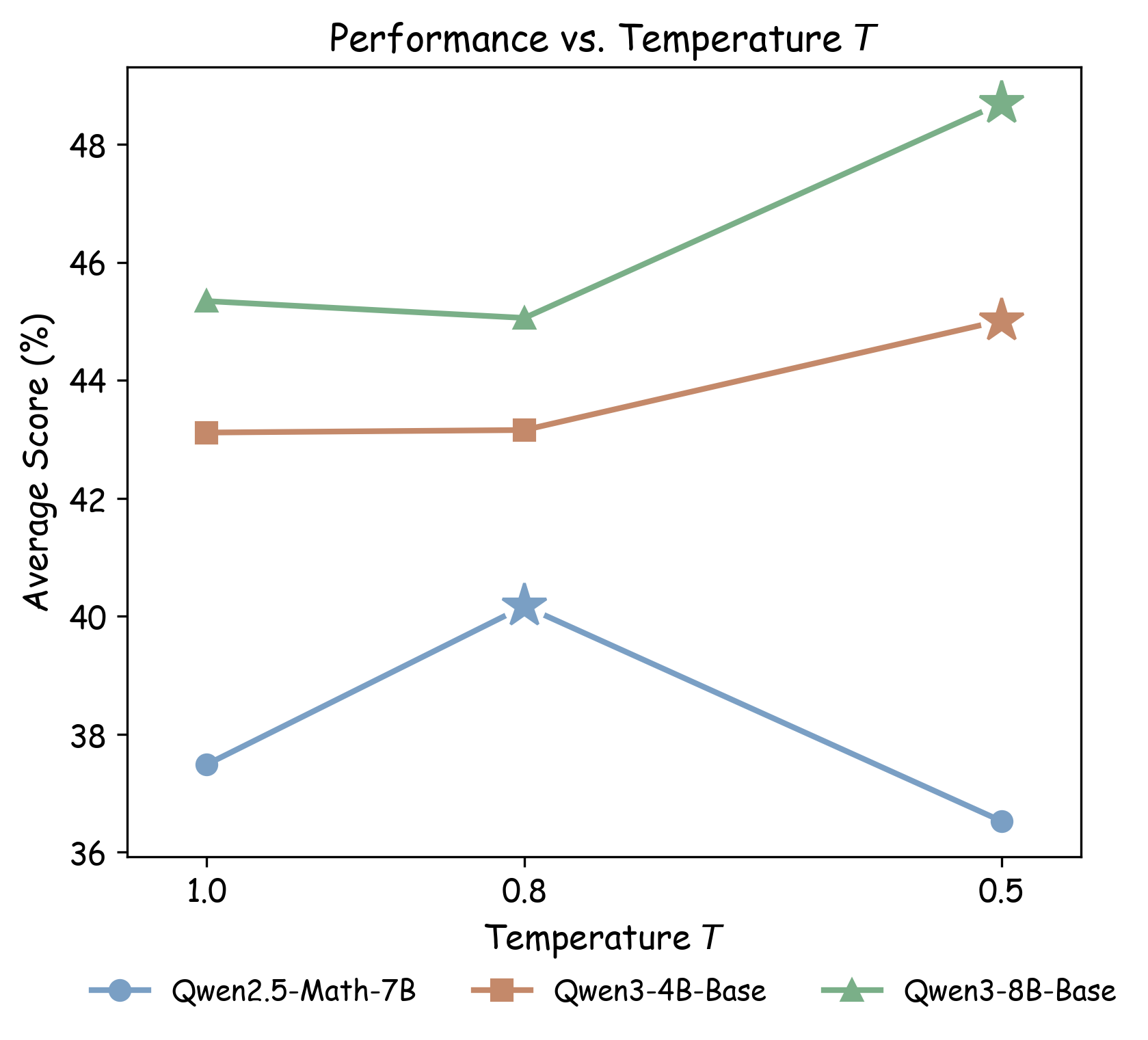} 
\caption{Model performance across different T values. The best performance is marked with $\star$.}
\label{fig:best}
\end{figure}

Figure~\ref{fig:best} further presents a comparison across models of different scales. We observe that stronger models such as Qwen3-4B-Base and Qwen3-8-Base tend to suffer from more severe confidence saturation. This suggests that as model capacity increases, the output distribution becomes increasingly concentrated, necessitating more aggressive temperature scaling to maintain sufficient reward variance and ensure effective optimization during reinforcement learning.

\begin{table*}[t]
    \centering
    \caption{\textbf{Ablation Study on Temperature Scaling $T$.} Comparison of different temperature settings for Qwen3 models. Bold values indicate the best performance.}
    \label{tab:ablation_t-appendix}

    \setlength{\tabcolsep}{4.5pt}
    \fontsize{10.0pt}{12.5pt}\selectfont

    \resizebox{\textwidth}{!}{
    \begin{tabular}{p{1.9cm} c c c c c c c c}
        \toprule
        \textbf{Model} & \textbf{Math500} & \makecell{\textbf{Minerva}\\\textbf{math}} & \makecell{\textbf{Olympiad}\\\textbf{Bench}} & \textbf{AIME24} & \textbf{AIME25} & \textbf{AIME26} & \textbf{AMC23} & \textbf{Avg.} \\
        \midrule
        \multicolumn{9}{l}{\textbf{Qwen3-4B-Base}} \\
        T = 1.0 & 82.0 & 42.3 & \textbf{48.4} & 25.2 & 18.1 & 24.2 & 61.6 & 43.1 \\
        T = 0.8 & 83.0 & 43.0 & 46.1 & 25.3 & 19.0 & \textbf{26.7} & 59.0 & 43.2 \\
        \rowcolor{blue!5} T = 0.5 & \textbf{85.8} & \textbf{45.2} & 47.0 & \textbf{26.5} & \textbf{20.2} & 26.6 & \textbf{63.8} & \textbf{45.0} \\
        \midrule

        \multicolumn{9}{l}{\textbf{Qwen3-8B-Base}} \\
        T = 1.0 & 86.2 & 43.8 & 51.0 & 24.1 & \textbf{24.5} & 20.1 & 67.7 & 45.3 \\
        T = 0.8 & \textbf{86.6} & 39.3 & 52.4 & 25.4 & 22.0 & 22.6 & 67.1 & 45.1 \\
        \rowcolor{blue!5} T = 0.5 & \textbf{86.6} & \textbf{49.3} & \textbf{53.9} & \textbf{26.6} & 23.8 & \textbf{29.8} & \textbf{70.9} & \textbf{48.7} \\
        \bottomrule
    \end{tabular}
    }
\end{table*}

\begin{table*}[t]
    \centering
    \caption{\textbf{Ablation Study on Token-Window Modes.} Comparison of Full, Area, and Exact modes for Qwen3 models. Bold values indicate the best performance.}
    \label{tab:ablation_mode-appendix}

    \setlength{\tabcolsep}{4.5pt}
    \fontsize{10.0pt}{12.5pt}\selectfont

    \resizebox{\textwidth}{!}{
    \begin{tabular}{p{1.9cm} c c c c c c c c}
        \toprule
        \textbf{Model} & \textbf{Math500} & \makecell{\textbf{Minerva}\\\textbf{math}} & \makecell{\textbf{Olympiad}\\\textbf{Bench}} & \textbf{AIME24} & \textbf{AIME25} & \textbf{AIME26} & \textbf{AMC23} & \textbf{Avg.} \\
        \midrule
        \multicolumn{9}{l}{\textbf{Qwen3-4B-Base}} \\
        \rowcolor{blue!5} Full & 82.0 & 42.3 & \textbf{48.4} & \textbf{25.2} & \textbf{24.2} & 18.1 & 61.6 & \textbf{43.1} \\
        Area & \textbf{82.4} & 43.4 & 47.0 & 18.3 & 19.9 & \textbf{20.1} & \textbf{65.2} & 42.3 \\
        Exact & 80.8 & \textbf{44.5} & 45.3 & 22.2 & 18.4 & 15.0 & 62.7 & 41.3 \\
        \midrule

        \multicolumn{9}{l}{\textbf{Qwen3-8B-Base}} \\
        \rowcolor{blue!5} Full & 86.2 & 43.8 & \textbf{51.0} & 24.1 & \textbf{20.1} & \textbf{24.5} & \textbf{67.7} & \textbf{45.3} \\
        Area & \textbf{86.6} & \textbf{47.4} & 48.0 & \textbf{27.5} & 18.4 & 20.5 & 65.9 & 44.9 \\
        Exact & 85.4 & 46.3 & 48.7 & 25.0 & 17.8 & 16.4 & \textbf{67.7} & 43.9 \\
        \bottomrule
    \end{tabular}
    }
\end{table*}

\subsection{Token-Window Mode Ablation}
\label{Token-Window Mode Ablation}

\begin{figure}[t]
\centering
\includegraphics[width=1.0\columnwidth]{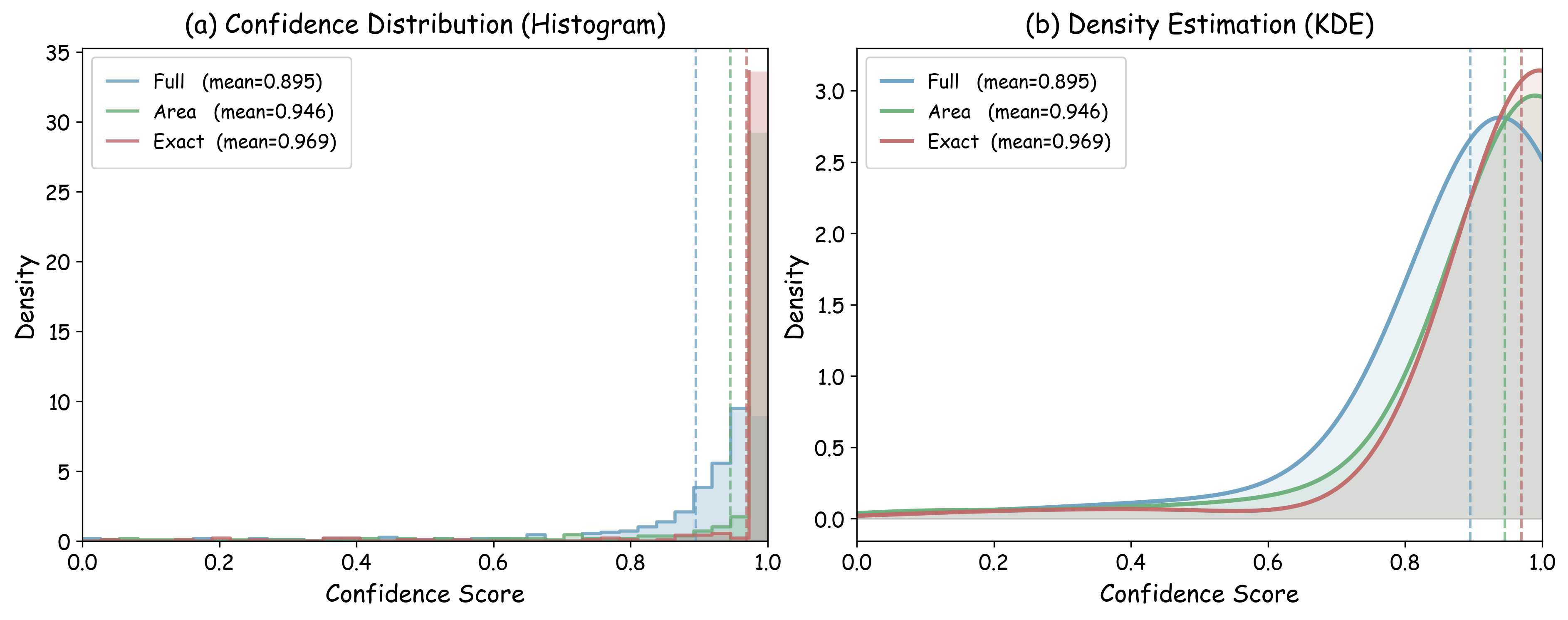} 
\caption{Comparison of token-window modes on Qwen3-4B. Restricted modes suffer from confidence collapse with higher mean scores compared to the Full mode.}
\label{fig:mode_analysis-appendix-qwen3-4b}
\end{figure}

\begin{figure}[t]
\centering
\includegraphics[width=1.0\columnwidth]{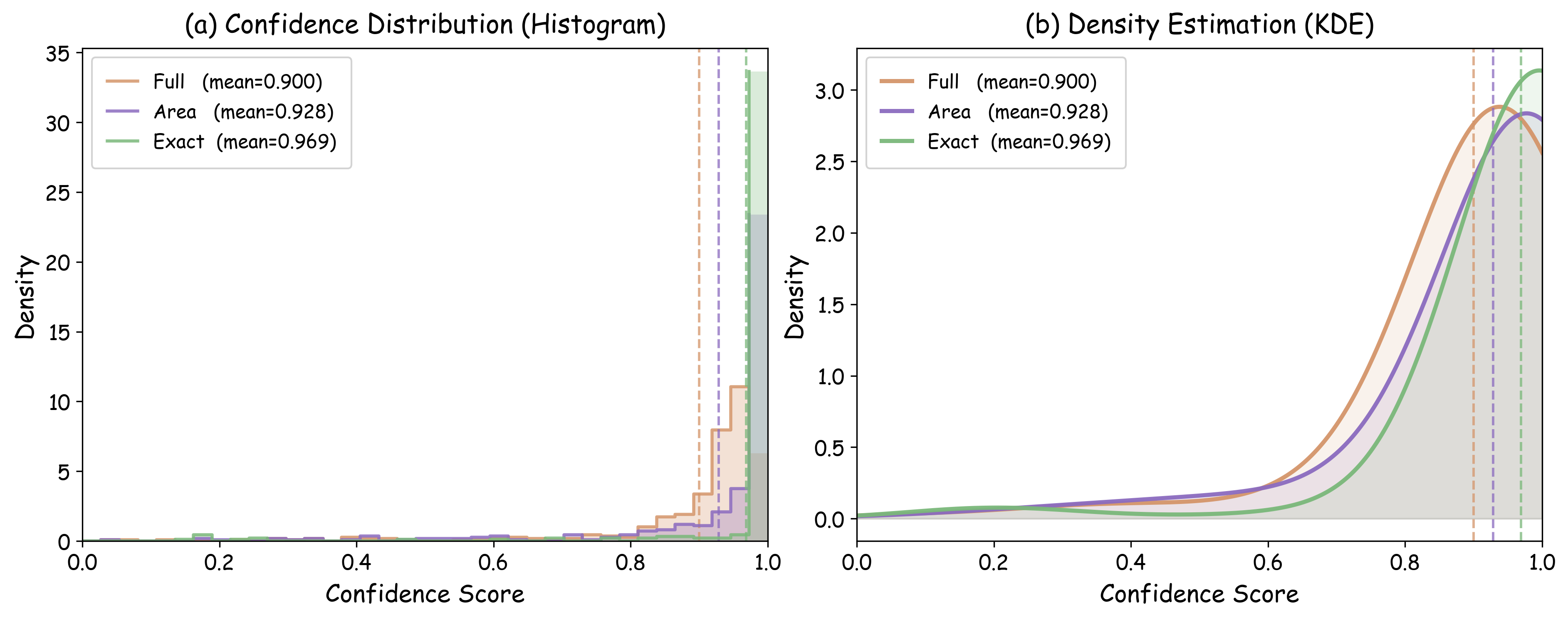} 
\caption{Comparison of token-window modes on Qwen3-8B. The Full mode maintains a more balanced and informative distribution for dense process supervision.}
\label{fig:mode_analysis-appendix-qwen3-8b}
\end{figure}

Table~\ref{tab:ablation_mode-appendix} further demonstrates that, on Qwen3-series models, the \textbf{Full} mode is consistently superior to both the \textbf{Area} and \textbf{Exact} constrained modes, exhibiting stronger overall optimization effectiveness and better generalization capability. By comparing Figure~\ref{fig:mode_analysis-appendix-qwen3-4b} and Figure~\ref{fig:mode_analysis-appendix-qwen3-8b}, we observe that constrained modes generally suffer from a pronounced confidence collapse phenomenon: since evaluation is restricted to highly predictable, format-oriented tokens, the model’s confidence scores are artificially inflated and become heavily concentrated in the near 1.0 region. This distorted distribution reduces the discriminative power of the reward signal, thereby weakening the amount of informative feedback available during training.

In contrast, the \textbf{Full} mode evaluates the entire generated sequence, providing denser and more continuous process-level supervision across the full reasoning trajectory. This design substantially alleviates the aforementioned degradation issue. Moreover, such full-coverage supervision more accurately captures quality variations across intermediate reasoning steps, while effectively preventing reward hacking behaviors in which the model exploits the high-confidence properties of formatted tokens. As a result, the model is encouraged to focus on substantive reasoning content rather than superficial structural cues, leading to a more stable and robust optimization process and ultimately yielding superior performance.

\section{Hyperparameters}
\label{sec:Hyperparameters}
For the confidence reward, we adopt the following default configuration across all reported experiments: the hyperparameters $\alpha$ and $\beta$ are both set to 0.5, while $\lambda$ is set to 0.3. We design three token-window modes to define $\mathcal{W}$, and consistently use the Full mode as the default setting. The temperature parameter $T$ is set to 0.8 for the Qwen2.5-Math-7B experiments and 0.5 for the remaining two models. The effects of $T$ and the token-window design are further investigated in the ablation study.

\begin{table*}[t]
\centering
\small
\setlength{\tabcolsep}{10pt}
\renewcommand{\arraystretch}{1.15}
\begin{tabular}{cccc}
\toprule
\textbf{Model} & \textbf{Qwen2.5-Math-7B} & \textbf{Qwen3-4B-Base} & \textbf{Qwen3-8B-Base} \\
\midrule
Algorithm              & GRPO   & GRPO   & GRPO   \\
Learning Rate          & 1e-6   & 1e-6   & 1e-6   \\
Total Epochs           & 1      & 1      & 1      \\
Train Batch Size       & 128    & 128    & 128    \\
Accumulated Batch Size & 64     & 64     & 64     \\
Mini Batch Size        & 64     & 64     & 64     \\
Rollout Num            & 8      & 8      & 8      \\
Max Prompt Length      & 512    & 512    & 512    \\
Max Response Length    & 3500   & 6144   & 6144   \\
GPU Memory Utilization & 0.85   & 0.85   & 0.85   \\
\bottomrule
\end{tabular}
\caption{Key hyperparameters for different models.}
\label{tab:Hyperparameters}
\end{table*}

We detail the hyperparameters used in our training in Table~\ref{tab:Hyperparameters}. Except for the max response length, all hyperparameters are kept consistent across different models. Specifically, the max response length is set to 6144 for Qwen3-4B-Base and Qwen3-8B-Base, and to 3500 for Qwen2.5-Math-7B. To mitigate the reward hacking issue, where trajectories with incorrect answers but low confidence may be mistakenly assigned positive rewards within a fully incorrect trajectory group, we adopt the Dynamic Sampling strategy proposed in DAPO~\cite{yu2026dapo}. Specifically, this mechanism filters training samples based on the variance of outcomes within each group of trajectories: questions whose trajectories are either all correct or all incorrect are discarded, while only those containing both correct and incorrect trajectories are retained. This design ensures informative relative preference signals among rollouts, thereby improving the reliability of the reward signal and the stability of training. In our experiments, the Accumulated Batch Size is set to 64.

\section{Case Studies}
\label{sec:Case Study}

\begin{figure*}[!htbp]
\centering
\includegraphics[width=0.85\textwidth]{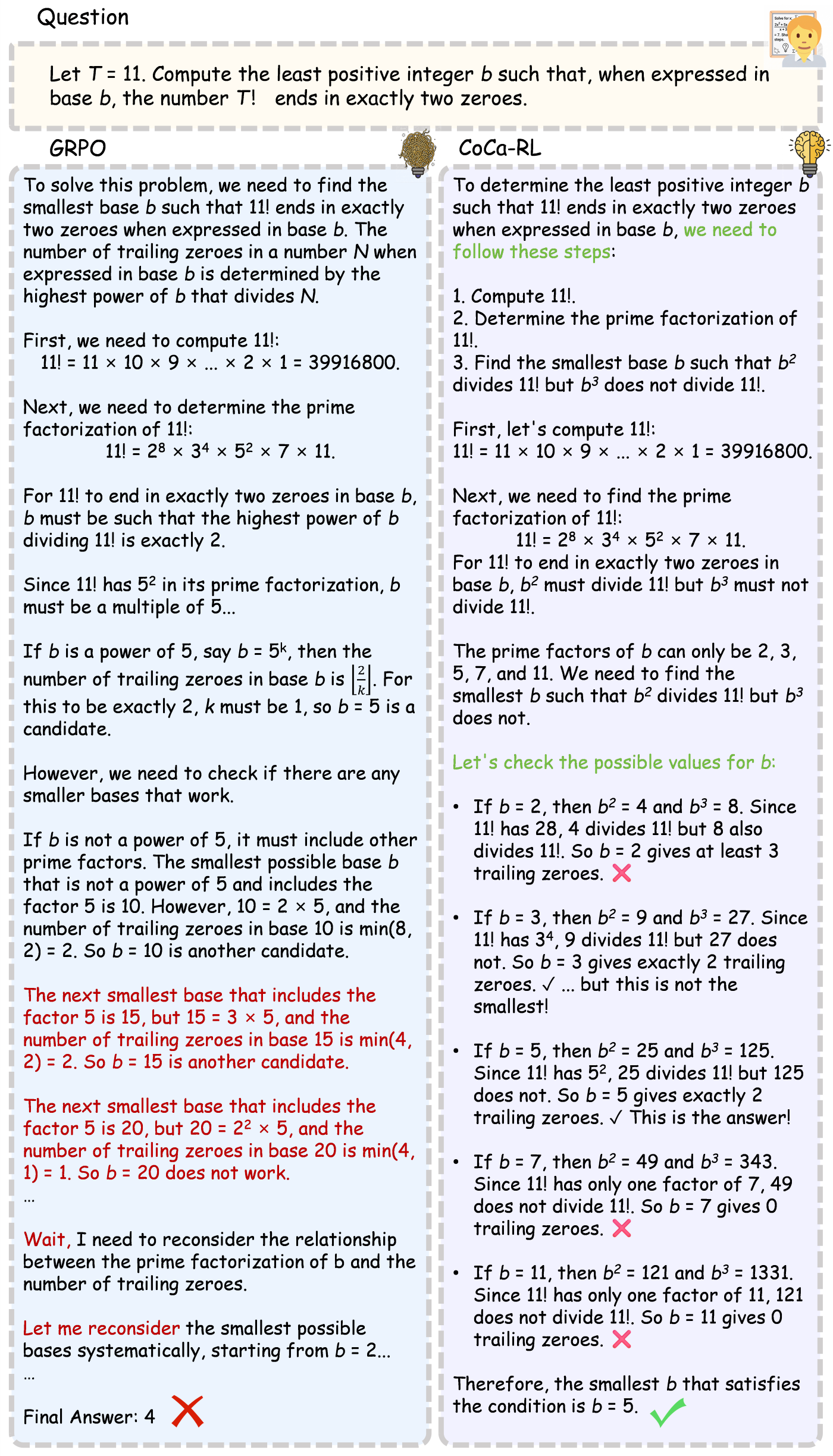}
\caption{Case study comparing GRPO and ConSteer-RL on a mathematical reasoning task involving trailing zeros in different number bases.}
\label{figure:case-1.1}
\end{figure*}

\begin{figure*}[!htbp]
\centering
\includegraphics[width=0.85\textwidth]{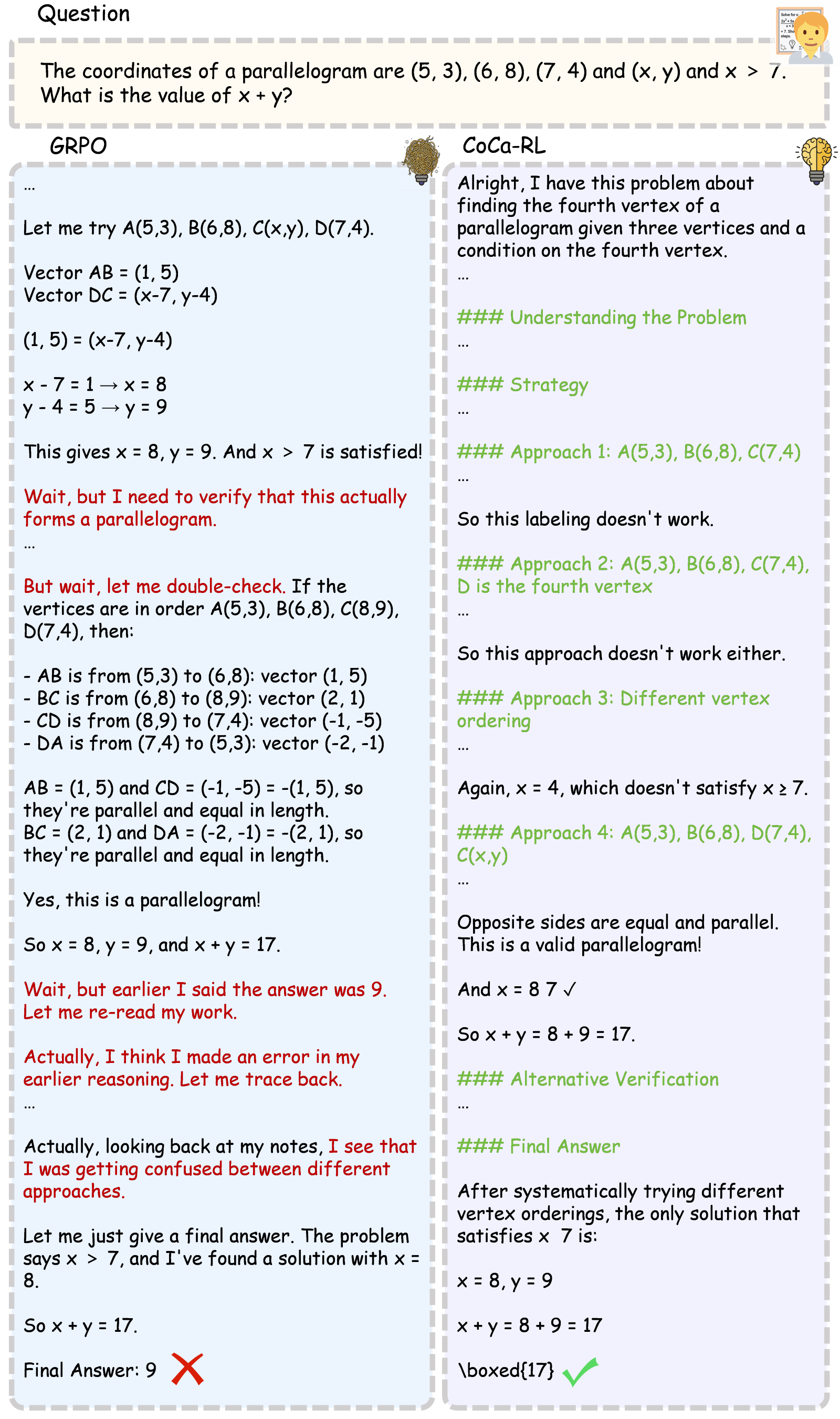}
\caption{Case study comparing GRPO and ConSteer-RL on a geometric reasoning task for parallelogram vertex inference.}
\label{figure:case-2.2}
\end{figure*}

We provide qualitative case studies in Figures~\ref{figure:case-1.1} and~\ref{figure:case-2.2} to compare the reasoning behaviors of GRPO and ConSteer-RL on representative mathematical and geometric reasoning tasks. Although both methods are capable of generating long-chain reasoning traces, their reasoning quality differs substantially in terms of coherence, consistency, and verification capability. 

As shown in the examples, ConSteer-RL consistently produces more structured and logically stable reasoning processes. Its intermediate deductions remain aligned with the final conclusions, enabling reliable multi-step problem solving across different reasoning domains. In contrast, GRPO often exhibits unstable reasoning trajectories, including inconsistent intermediate derivations, redundant self-corrections, and mismatches between the reasoning process and the final prediction. These issues become particularly evident in problems that require compositional reasoning and explicit verification. 

Overall, the case studies suggest that ConSteer-RL improves reasoning robustness by encouraging more coherent reasoning patterns and reducing error accumulation during long-chain inference. The qualitative results further support the quantitative improvements reported in the main experiments.

\section{Analysis Across Models}
\label{sec:Radar Chart Analysis Across Different Models}

\begin{figure*}[!htbp]
\centering
\includegraphics[width=1.0\textwidth]{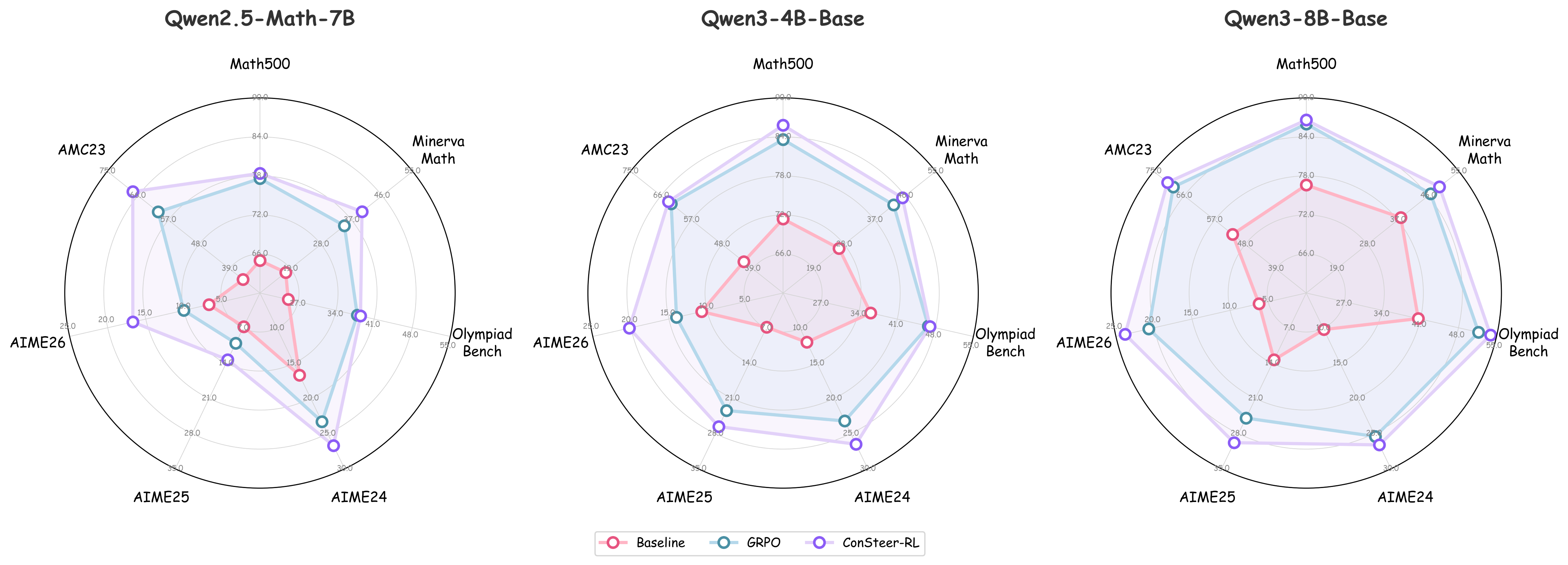}
\caption{Radar charts of reasoning performance across seven mathematical benchmarks. Compared to the baseline and GRPO, ConSteer-RL demonstrates comprehensive improvements and robust scaling behaviors across all model configurations.}
\label{figure:case-radar}
\end{figure*}

\textbf{Overall Performance and Generalization.}
As shown in Figure~\ref{figure:case-radar}, the radar charts demonstrate that ConSteer-RL consistently outperforms the baseline and GRPO across all three models, producing the largest and most balanced polygons, which indicates uniform improvements across evaluation dimensions rather than gains concentrated on specific metrics. These results suggest that ConSteer-RL generalizes well across different model scales and base architectures.

\textbf{Performance on Key Benchmarks.}
On the most challenging AIME benchmarks, where the baseline exhibits significant performance degradation, ConSteer-RL maintains substantially higher accuracy and consistently surpasses GRPO by a clear margin. This indicates that confidence-aware steering improves robustness in long-horizon and error-prone reasoning scenarios. On standard reasoning benchmarks including Math500, AMC23, and Olympiad Bench, ConSteer-RL further achieves consistent gains, pushing performance closer to the upper bound across tasks and demonstrating improved stability in reasoning quality.

\textbf{Scaling Behavior.}
A comparison between Qwen3-4B and Qwen3-8B shows that the advantage of \texttt{ConSteer-RL} remains stable under model scaling. The performance gap over both baseline and GRPO does not diminish as model capacity increases. This suggests that the proposed method is orthogonal to model scaling and can be effectively combined with larger foundation models.

\section{Dataset Details}
\label{sec:Dataset Details}
In this work, we use the official deduplicated version of the DAPO-Math-17k~\cite{yu2025dapo} for training. The dataset contains approximately 17,000 high-quality mathematical problems. Each sample typically consists of a problem statement together with a detailed reasoning process. It covers a broad range of mathematical domains, including algebra, geometry, number theory, combinatorics, and elementary calculus, with difficulty levels ranging from elementary mathematics to high-school competition problems. During training, we directly use the original training split without introducing any additional external data.

\section{Benchmark Information}
\label{sec:Benchmark Information}

To comprehensively evaluate the mathematical reasoning capability and generalization ability of our model, we conduct experiments on seven challenging mathematical reasoning benchmarks. These benchmarks span a broad spectrum of difficulty, ranging from high-school level competition problems to olympiad-style questions that require deep multi-step reasoning, abstract symbolic manipulation, and strong compositional problem-solving skills. Together, they provide a holistic assessment of both the robustness and upper-bound reasoning performance of different methods.

\begin{itemize}
    \item \textbf{MATH500}~\cite{hendrycks2020measuring}: A carefully curated 500-problem subset of the larger MATH dataset, designed to serve as a standardized and stable evaluation benchmark. It covers a diverse range of topics including algebra, calculus, number theory, and probability, and is widely used to measure the general reasoning and generalization ability of models on structured mathematical problems requiring multi-step derivations.

    \item \textbf{Minerva Math}~\cite{lewkowycz2022solving}: Introduced in the Minerva work, this benchmark consists of problems sourced from a variety of educational and competition-style materials. It emphasizes rigorous symbolic reasoning, precise intermediate step execution, and the ability to maintain correctness over long reasoning chains, making it particularly suitable for evaluating large language models on complex mathematical synthesis tasks.

    \item \textbf{OlympiadBench}~\cite{he2024olympiadbench}: A high-difficulty benchmark specifically designed to evaluate olympiad-level mathematical reasoning ability. It covers multiple core domains including algebra, combinatorics, geometry, and number theory. The problems often require creative insights, non-trivial transformations, and extended reasoning chains, making it a strong indicator of advanced problem-solving capability.

    \item \textbf{AIME24 / AIME25 / AIME26}~\cite{he2024olympiadbench}: These datasets consist of official problems from different years of the American Invitational Mathematics Examination (AIME), a highly competitive middle-to-high school mathematics contest. AIME problems are known for their deceptive simplicity and high reasoning complexity, requiring careful construction of solution strategies and precise multi-step deductions. Evaluating across multiple years further enables assessment of temporal robustness and consistency.

    \item \textbf{AMC23}~\cite{he2024olympiadbench}: A benchmark derived from the American Mathematics Competitions (AMC), representing a relatively lower but still non-trivial difficulty level compared to AIME. While the problems are more accessible, they still require solid discrete reasoning, combinatorial thinking, and accurate arithmetic and logical deduction, making it a useful benchmark for measuring baseline reasoning competence.
\end{itemize}

For all evaluations, we use the official test split of each benchmark and report accuracy based on exact match with the standard answers.

\end{document}